\documentclass[a4paper,twoside]{article}

\usepackage{epsfig}
\usepackage{subcaption}
\usepackage{calc}
\usepackage{amssymb}
\usepackage{amstext}
\usepackage{amsmath}
\usepackage{amsthm}
\usepackage{multicol}
\usepackage{pslatex}
\usepackage{apalike}
\usepackage{algorithm2e}
\usepackage{url}

\usepackage[bottom]{footmisc}
\usepackage{SCITEPRESS}     

\begin{document}
\onecolumn
\noindent Accepted pre-print version.\\ \\\\\\ Cite as: \\\\ Balestri, R. and Pescatore, G. (2025).\\\\
\textit{Multi-Agent System for AI-Assisted Extraction of Narrative Arcs in TV Series.}\\\\
In Proceedings of the 17th International Conference on Agents and Artificial Intelligence - Volume 1,\\\\
ISBN 978-989-758-737-5, ISSN 2184-433X, pages 663-670. \\\\DOI: 10.5220/0013369600003890\\\\
\url{https://www.scitepress.org/PublicationsDetail.aspx?ID=nm4GOKdsrzg=&t=1}\\\\
Under Creative Commons license CC BY-NC-ND

\title{Multi-Agent System for AI-Assisted Extraction of Narrative Arcs in TV Series}

\author{\authorname{Roberto Balestri\sup{1}\orcidAuthor{0009-0000-5008-2911}, Guglielmo Pescatore\sup{1}\orcidAuthor{0000-0001-5206-6464}}
\affiliation{\sup{1}Department of the Arts, Università di Bologna, Italy}
\email{roberto.balestri2@unibo.it, guglielmo.pescatore@unibo.it}
}

\keywords{Multi-Agent Systems, Narrative Analysis, TV Series, Computational Narratology, LLM}

\abstract{Serialized TV shows are built on complex storylines that can be hard to track and evolve in ways that defy straightforward analysis. This paper introduces a multi-agent system designed to extract and analyze these narrative arcs. Tested on the first season of \textit{Grey's Anatomy} (ABC 2005-), the system identifies three types of arcs: Anthology (self-contained), Soap (relationship-focused), and Genre-Specific (strictly related to the series' genre). Episodic progressions of these arcs are stored in both relational and semantic (vectorial) databases, enabling structured analysis and comparison.
To bridge the gap between automation and critical interpretation, the system is paired with a graphical interface that allows for human refinement using tools to enhance and visualize the data. The system performed strongly in identifying Anthology Arcs and character entities, but its reliance on textual paratexts (such as episode summaries) revealed limitations in recognizing overlapping arcs and subtler dynamics.
This approach highlights the potential of combining computational and human expertise in narrative analysis. Beyond television, it offers promise for serialized written formats, where the narrative resides entirely in the text. Future work will explore the integration of multimodal inputs, such as dialogue and visuals, and expand testing across a wider range of genres to refine the system further.}

\onecolumn \maketitle \normalsize \setcounter{footnote}{0} \vfill

\section{\uppercase{Introduction}}

\label{sec:introduction}

The analysis of narrative structure has long been a focus of inquiry in both literary studies and media analysis. From Formalism and Structuralism to the more contemporary methodologies of Classical and Post-Classical Narratology, the study of narrative has centered on understanding the underlying relationships and components that constitute storytelling \cite{ionescu2019postclassical}. The process of narrative understanding mirrors other forms of comprehension: it involves identifying constituent elements and integrating them into a coherent whole.

 Television series pose unique challenges, with storylines interweaving over seasons, resisting straightforward categorization \cite{mittell2015complex}. These "narrative arcs" are not singular or linear but distributed across episodes and interconnected with other storylines. Some arcs conclude within an episode, while others develop gradually, spanning seasons. Understanding these patterns requires approaches that can identify connections and developments across diverse temporal scales.

 This paper introduces a multi-agent system designed to assist in the extraction and mapping of narrative arcs from serialized television content. The system parses episode summaries to identify arcs and their progressions, categorizing them into types such as self-contained, interpersonal, or genre-specific arcs. These arcs are stored in a relational database and enriched with semantic embeddings in a vector database. To test the system, we analyzed the first season of \textit{Grey's Anatomy} (ABC 2005-), comparing the extracted arcs to those identified by a human scholar familiar with the series. 

 The remainder of this paper is organized as follows. Section~\ref{sec:related_works} situates this study within the broader context of narrative analysis and computational methods. Section~\ref{sec:narrative_arcs} introduces the conceptual framework for modeling and storing narrative arcs. Section~\ref{sec:technical} describes the technical infrastructure supporting the system. Section~\ref{sec:material_collection} outlines the data collection and preprocessing pipeline of the episodes' plots. Section~\ref{sec:mas_extraction} presents the multi-agent framework for extracting and organizing narrative arcs. Section~\ref{sec:graphic} explores the graphical interface designed to facilitate human oversight and refinement of the system’s outputs. Section~\ref{sec:test_improv} evaluates the system’s performance, discusses its strengths and limitations, while \ref{sec:conclusion} closes the paper identifying potential improvement.

 The GitHub repository containing the code and the instructions to run the software are available at: https://github.com/robertobalestri/MAS-AI-Assisted-Narrative-Arcs-Extraction-TV-Series .

\section{\uppercase{Related Works}}
\label{sec:related_works}

The study of narrative structures and their evolution has been a foundational area of research across disciplines, from literary theory to computational linguistics. Movements such as Formalism and Structuralism in the 20th century explored the mechanics of storytelling, emphasizing recurring patterns and structures. Pioneering works like Propp’s analysis of Russian folktales \cite{propp2012russian} and Todorov’s structural analysis \cite{todorov1969structural} established frameworks that continue to inform both theoretical and computational methods for narrative understanding. In recent years, advances in natural language processing (NLP) and machine learning have expanded the scope of computational narrative studies \cite{piper2021narrative}.

\subsection{Advances in Computational Narrative Analysis}

Early computational methods, including Support Vector Machines \cite{hearst1998support} and logistic regression \cite{vimal2020application}, struggled to capture the complex relationships and temporal dynamics inherent in narratives \cite{young2018recent}. The advent of deep learning, particularly recurrent neural networks (RNNs) and convolutional neural networks (CNNs), marked significant progress by enabling the modeling of sequential and local features in text \cite{yin2017comparative}.

Transformer-based models, such as GPT \cite{radford2018improving} and BERT \cite{vaswani2017attention}, further revolutionized the field with self-attention mechanisms that capture long-range text dependencies. These models leverage embeddings to represent semantic relationships mathematically, enabling deeper narrative understanding and generation \cite{haywood2024tokens}. More recently, Retrieval-Augmented Generation (RAG) has improved narrative interpretation by integrating external knowledge sources \cite{lewis2020retrieval}.

\subsection{Temporal and Structural Analysis of Narratives}
Methods like sentiment analysis and topic modeling have been used to identify emotional arcs and thematic progressions in narratives \cite{reagan2016emotional, schmidt2015plot}. However, advanced models are needed to address non-linear temporal dynamics and complex story turning points \cite{piper2023quantitative}. Concepts such as narrative flow, suspense, and causality remain underexplored despite their importance for audience engagement \cite{wilmot-keller-2020-modelling}.

Television series, with their layered and evolving storylines, pose unique challenges for narrative analysis. Unlike standalone stories, TV series often feature arcs spanning multiple episodes or seasons, requiring robust methods to identify and track these connections \cite{mittell2015complex}. While NLP techniques have been employed to extract narrative elements and character interactions \cite{dalla20233,janosov2021network,bost2016narrative}, these methods often rely on significant manual effort \cite{rocchi2022}.

\subsection{Multi-Agent Systems in Narrative Understanding}

Large language models (LLMs) have shown potential in narrative generation and analysis \cite{zhao2023narrativeplay}. Multi-agent systems (MAS) further enhance these capabilities by enabling collaborative task processing. MAS have been used to advance fields such as legal reasoning \cite{yuan2024can} and scientific theory development \cite{ghafarollahi2024sciagents}, as well as narrative generation \cite{aoki2023analysis}.

By assigning specialized tasks to individual agents, MAS facilitate the extraction of narrative arcs and the tracking of their progression across episodes, making them well-suited for analyzing serialized narratives.

\subsection{Narrative Arcs (Plotlines)}
TV series diverge significantly from the classical narration model, relying on networks of interconnected storylines rather than single linear plots \cite{perez2021multi}. Serialized narratives operate through a multi-layered structure that evolves across episodes and seasons, creating a unique sense of depth and continuity \cite{pescatore2019narration,rocchi2022,esposti2023exploring}.

Narrative arcs in TV series can be categorized into Anthology Plots, which resolve within a single episode, and Running Plots, which span multiple episodes or seasons. Running Plots include Soap Plots, focusing on character relationships, and Genre-Specific Plots, tied to the show’s thematic or professional elements, such as survival challenges in a zombie series.

These categories build on the framework outlined in \cite{pescatore2019narration}, which uses the concept of isotopy to ensure textual coherence \cite{greimas1982semiotics}. Episodic isotopies define Anthology Plots, interpersonal and intrapersonal isotopies sustain Soap Plots, and thematic isotopies characterize Genre-Specific Plots.

To generalize this framework beyond medical dramas, where Running Plots were initially defined as Sentimental and Professional Plots \cite{pescatore2019narration}, we propose broader definitions applicable to other serialized genres.

\section{\uppercase{Modeling Narrative Arcs for Storing}}
\label{sec:narrative_arcs}
To analyze and document these arcs effectively, we conceptualized a model that captures their key attributes and episodic progressions in a structured format.
Each arc is categorized into one of three types based on its nature and scope:
\begin{itemize}

    \item Anthology Arcs, which are self-contained stories resolved within a single episode, often centered on genre-specific events or cases.
    \item Genre-Specific Arcs, focusing on professional or thematic elements tied to the series, such as workplace dynamics or medical conflicts.
    \item Soap Arcs, which explore interpersonal relationships, personal growth, and emotional conflicts.
\end{itemize}

 Each arc is defined by a title and description that encapsulates its central theme or conflict. The main characters driving the arc are identified, along with the arc type, which provides context for its role within the series.
 To track how arcs evolve over time, we document their progressions, which capture episodic developments relevant to each storyline. A progression specifies the episode and season where it occurs, the interfering characters influencing the arc in that context, and a concise description of the events advancing the storyline.

 Summarizing, in our system, the object \textit{Narrative Arc} includes several fields. Each arc is assigned a unique identifier (\textit{arc\_id}), along with a title and description encapsulating its central theme or conflict. The arcs also contain progressions, which are a list of \textit{Progression} objects representing their developments across episodes. Additionally, each arc identifies the main characters driving the storyline, specifies the arc type (Anthology, Soap, or Genre-Specific), and indicates the series to which it belongs.

Each \textit{Progression} is similarly structured with key attributes. It is assigned a unique identifier (\textit{progression\_id}) and is linked to its corresponding narrative arc through the \textit{arc\_id}. The progression includes a description of the events related to the arc in the specific episode (\textit{Content}), as well as information about the series, season, and episode in which it occurs.

\section{\uppercase{Technical Details}}
\label{sec:technical}
In this paper, "LLM" refers specifically to the OpenAI GPT-4o model, accessed via API. Data storage was implemented using an SQLite database, while embeddings were generated with the Cohere-embed-english-v3.0 model and stored in a Chroma vector database. The backend was developed in Python (version 3.11), with FastAPI facilitating communication between the backend and frontend, which was built using the React JavaScript framework. Character entities were extracted from episode plots using the Spacy NLP model \textit{en\_core\_web\_trf}.

\section{\uppercase{Material Collection and Preprocessing}}
\label{sec:material_collection}

Our software was designed to be generalizable across various TV series genres. For testing, we focused on the first season of \textit{Grey's Anatomy}, a choice motivated by our team’s prior research on medical dramas. This familiarity provided a solid foundation for evaluating the software’s performance. The system requires only episode summaries of a season to function effectively.

\subsection{Episodes' Plots Gathering}
To develop and test the software, we sourced episode summaries from the fan-maintained \textit{Grey's Anatomy Wiki} (\url{https://greysanatomy.fandom.com}), which operates under the Creative Commons Attribution-Share Alike License (CC BY-SA). This license permits research and commercial use with proper attribution. The raw data was preprocessed to ensure standardization and usability for analysis.

\subsection{Data Cleaning and Simplification}
The preprocessing began with cleaning and simplifying the episode summaries. Using the LLM, sentences were rewritten in simpler, more structured forms, emphasizing clarity by focusing on a single character or event per sentence. Direct quotations were avoided, and complex sentence structures were replaced with concise descriptions. This process ensured the plots were easily interpretable and consistent.

\subsection{Character Entity Normalization}
Character name variability was addressed by replacing pronouns with corresponding character names, guided by contextual analysis within a fifteen-sentence window. The spaCy Transformer model was used to extract character entities, and the LLM further refined this output by resolving similar names and alternative appellations. A character database was created, containing unique identifiers, preferred names, and alternative names for each character. The plots were then standardized by replacing all character references with their preferred names.

\subsection{Season Plot Generation}
To create a cohesive season-level summary, individual episode plots were first summarized using the LLM. These summaries were then concatenated and further summarized to produce a streamlined narrative overview of the season. This two-step process ensured the essential details of the narrative were retained while providing a comprehensive summary.

This preprocessing pipeline transformed raw textual data into a structured format, enabling detailed narrative analysis and ensuring the software’s applicability across various TV series genres.

\section{\uppercase{The Multi-Agent System for Arc Extraction}}
\label{sec:mas_extraction}

\begin{figure}
\centering
\includegraphics[width=1\linewidth]{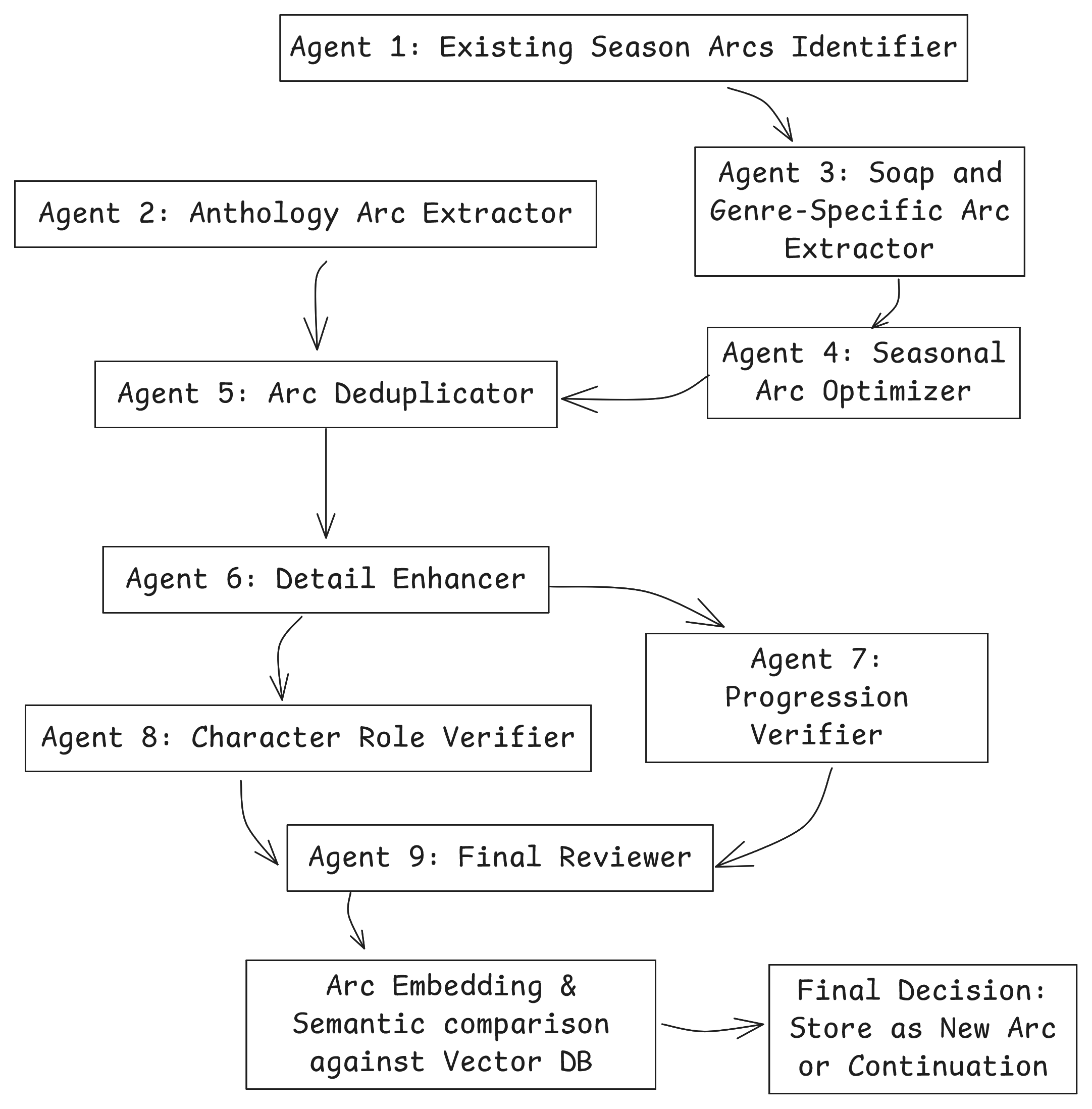}
\caption{Narrative Arc Extraction Process}
\label{fig:extraction}
\end{figure}

The narrative arc extraction process utilizes a multi-agent system to analyze the episode plots. It identifies, categorizes, and refines narrative arcs, integrating episodic storylines into a cohesive seasonal framework. The workflow is sequential, with specialized agents performing tasks. The resulting arcs and their progressions are stored in both a relational database and a vector database, as detailed in Section \ref{sec:semantic_comparison}. Figure \ref{fig:extraction} provides a visual representation of the system.

\subsection{Workflow Design}
The extraction workflow is divided into stages, with each stage handled by an autonomous agent. Each agent focuses on a specific aspect of narrative analysis, ensuring arcs are both episode-specific and seasonally coherent. Results from prior episodes are incorporated to maintain continuity and account for the evolving nature of serialized storytelling.

\subsubsection{Agent 1 - Existing Season Arcs Identifier}
This agent evaluates if arcs from previous episodes (if present) are present in the current episode. If detected, they are flagged as "possibly present" and serve as reference points for subsequent agents.

\subsubsection{Agent 2 - Anthology Arc Extractor}
This agent identifies self-contained, standalone storylines unique to the current episode. These arcs are processed independently, as they do not contribute to season-wide continuity.

\subsubsection{Agent 3 - Soap and Genre-Specific Arc Extractor}
This agent analyzes the episode plot to identify new soap and genre-specific arcs and to validate arcs flagged by Agent 1.

\subsubsection{Agent 4 - Seasonal Arc Optimizer}
This agent minimizes redundancy by analyzing soap and genre-specific arcs for overlaps. It performs stricter checks on "possibly present" season arcs and newly identified arcs, merging or refining them as needed to maintain distinct scopes.

\subsubsection{Agent 5 - Arc Deduplicator}
All arcs extracted from the episode, including Anthology Arcs, are reviewed for similarity. Overlapping arcs are resolved through a disambiguation process. For example, arcs identified as both Anthology and Genre-Specific by separate agents are clarified by this deduplicator.

\subsubsection{Agent 6 - Detail Enhancer}
This agent enriches each arc with detailed contextual information, including: main characters driving the arc, supporting or interfering characters influencing the arc in the episode, a concise description of the arc’s episodic events (the "progression").

\subsubsection{Agent 7 - Progression Verifier}
The progressions of each arc are reviewed to ensure specificity and relevance. This step ensures that significant developments are captured without overlapping with unrelated narratives.

\subsubsection{Agent 8 - Character Role Verifier}
This agent verifies and, if necessary, corrects the classification of characters as either main or interfering.

\subsubsection{Agent 9 - Final Reviewer}
A final verification ensures narrative consistency. Validated arcs are stored in the database for future episodes' analysis.

\subsection{Semantic Comparison and Arc Embedding}
\label{sec:semantic_comparison}
After analyzing each episode, embeddings are generated for the arcs. The embeddings represent the "Title" and "Description" of each arc, while the "Content" of the progression provides additional context.

These embeddings are compared against the vector database to identify semantically similar arcs. If similarities are detected, an LLM determines whether the arcs represent the same storyline or not. Based on this determination, the system either stores them as new arcs or links them as continuations of overarching plotlines.

\section{\uppercase{Graphic Interface}}
\label{sec:graphic}

Human input remains essential for refining the outputs of the multi-agent system. The graphical interface is designed to facilitate this oversight and correction, balancing simplicity with advanced features. It enables users to visualize, edit, regenerate, and analyze narrative arcs effectively.

\subsection{Interface Overview}

The interface is divided into three main sections:
\begin{itemize}
\item Narrative Arc Timeline: Provides a visual representation of how storylines evolve across episodes in a season. Users can apply filters based on arc type (Anthology, Genre-Specific, Soap) or associated characters.
\item Vector Store Explorer: Includes tools for visualizing arc embeddings using 3D PCA and displays clusters of similar arcs.
\item Character Section: Allows users to explore extracted character entities, manage appellations, or merge similar characters.
\end{itemize}

\subsection{Key Features and Functionalities}

\subsubsection{Arc and Progression Visualization}
\begin{figure}
\centering
\includegraphics[width=0.99\linewidth]{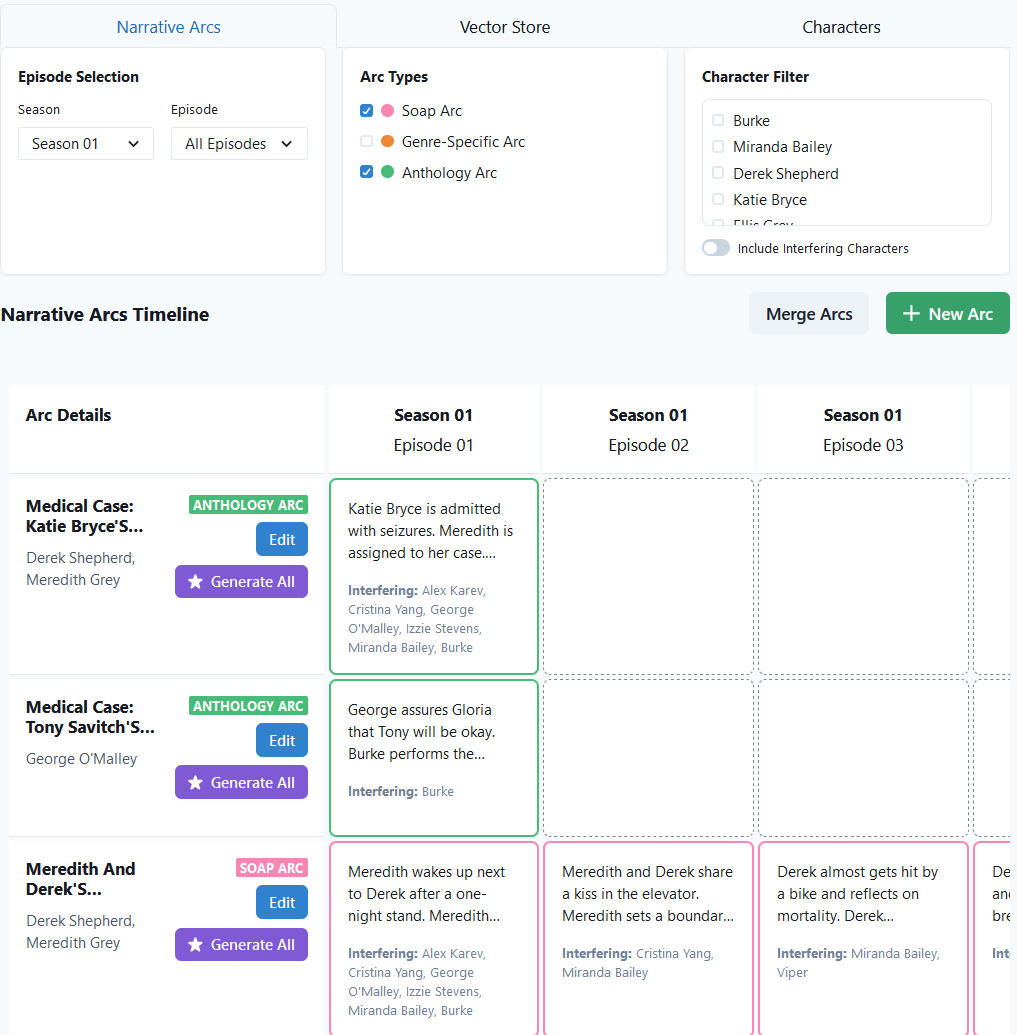}
\caption{The main view of the graphical interface.}
\label{fig:main_view}
\end{figure}

Users can navigate narrative arcs through a tabular interface, where rows represent arcs, and columns correspond to episodes. Each cell displays the arc’s progression within a specific episode (see Figure \ref{fig:main_view}).

\subsubsection{Arc Creation and Editing}
\begin{figure}
    \centering
    \includegraphics[width=0.75\linewidth]{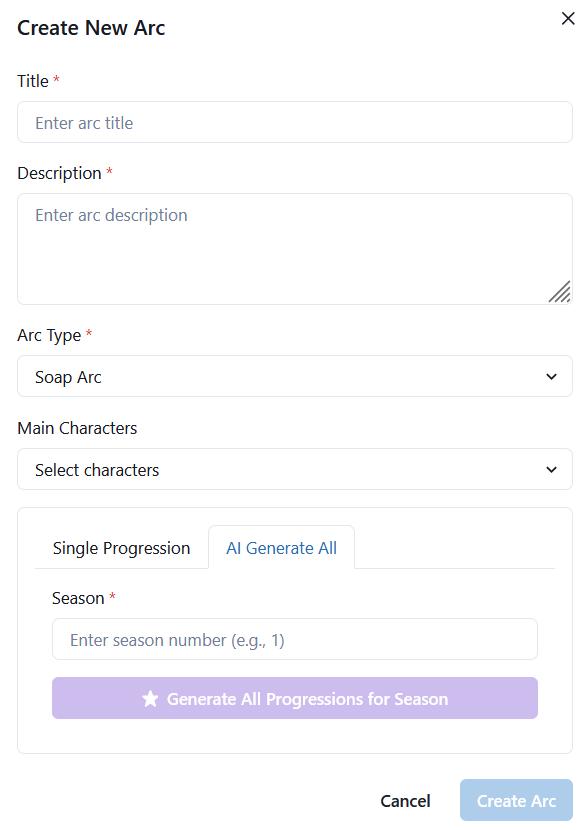}
    \caption{Form to add a new arc.}
    \label{fig:new_arc}
\end{figure}

Arcs can be created or edited via a dialog box (see Figure \ref{fig:new_arc}). Users specify attributes such as title, description, and arc type, as well as main characters and episodic progressions. Progressions can also be auto-generated using AI for efficiency.

\subsubsection{Arc Merging and Deduplication}
For overlapping or semantically similar arcs, users can compare them side-by-side and merge duplicates to maintain consistency.

\subsubsection{Progression Management}
Users can manually create or edit episodic progressions, which include episode-specific descriptions and interfering characters. Alternatively, progressions can be auto-generated and refined as needed.

\subsubsection{Clustering and Semantic Analysis}
\begin{figure}
\centering
\includegraphics[width=0.99\linewidth]{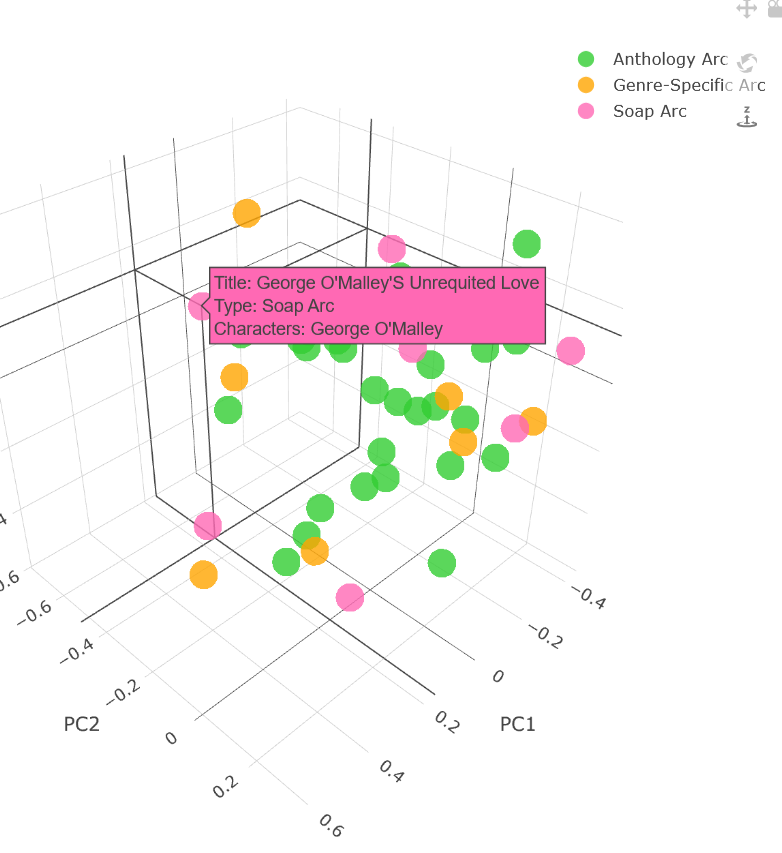}
\caption{3D PCA visualizer for clustering arcs based on semantic similarity.}
\label{fig:3d_pca}
\end{figure}

Clustering tools group arcs based on semantic embeddings and display them in both tabular and 3D PCA formats (see Figure \ref{fig:3d_pca}). This feature helps users identify thematic connections and detect arcs that the multi-agent system incorrectly treated as distinct.

\subsubsection{Character Management}
\begin{figure}
\centering
\includegraphics[width=0.75\linewidth]{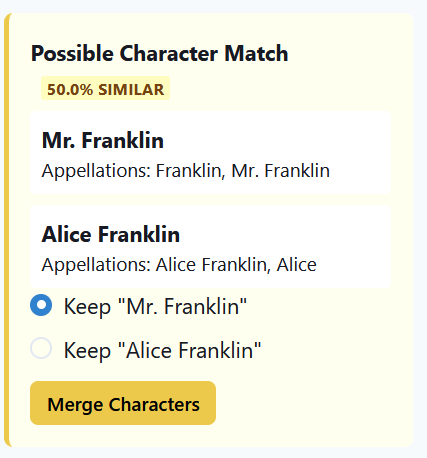}
\caption{Jaccard Index indicating potential duplicated characters. In this example, the suggestion is a false positive.}
\label{fig:character_merge}
\end{figure}

The character panel enables users to view, edit, and merge characters to ensure consistency across arcs. A similarity threshold based on the Jaccard index \cite{niwattanakul2013using} highlights potential duplicate characters for resolution. For example, Figure \ref{fig:character_merge} illustrates a false positive where two different characters with the same surname are identified as similar by the system.

\section{\uppercase{Testing the System}}
\label{sec:test_improv}

\subsection{Test Setup}

The system’s performance was evaluated by comparing its extracted narrative arcs, based solely on episode summaries (paratexts), with arcs identified by a human scholar who analyzed the first season of \textit{Grey's Anatomy} after watching each episode at least twice.

Paratexts \cite{genette1997paratexts}, such as episode plots, fan fiction, plot maps, and similar materials, have gained increasing significance in the Internet era \cite{laukkanen2024audience}. While these materials do not, by their intrinsic nature, constitute the primary text itself, they provide scholars like us with valuable supplementary resources for research and analysis.

The test was conducted across the entire season, and the results were analyzed to assess both the system's capabilities and its limitations.

\subsection{Results}

The system demonstrated notable strength in its ability to perform LLM-aided character entity recognition and linking. It successfully identified 62 character entities within the test dataset, with 61 of these being correct. The single duplicate entry ("Frost" and "Jerry Frost") resulted from inconsistent naming in the source material. Implementing a two-agent verification system could resolve such errors but would increase computational costs.

In arc extraction, the system excelled in identifying Anthology Arcs, achieving a precision of 89.3\% (25 out of 28 arcs correctly extracted).

However, challenges were observed with other arc types. The system occasionally duplicated arcs or failed to merge arcs representing the same storyline. For example, the arcs:

\begin{itemize} \item "Izzie Stevens: Overcoming Past and Professional Growth" \item "Izzie Stevens: Balancing Personal Life and Professional Ambitions" \end{itemize}

were treated as separate, whereas human analysis considered them part of a unified storyline.

Conversely, the system overlooked the shared arc of Meredith Grey and Derek Shepherd’s relationship. Instead, it identified only individual character arcs for Meredith and Derek, effectively "diluting" their shared storyline within the personal arcs of each character.

Additionally, the system misclassified the "Roommates Dynamics" arc, which involves Meredith, Izzie, and George becoming roommates in the second episode. This storyline was grouped under a broader arc, "Intern Dynamics: Friendship And Rivalry," though human analysis deemed them distinct.

Progressions within arcs were generally consistent with their titles and descriptions. However, they were not always fully captured for specific episodes, as episode summaries often omit minor developments that a viewer might identify.

\subsection{Impact of the Graphic Interface}

The graphical interface proved highly effective in refining the system's outputs. It enabled human analysts to review, correct, and enhance extracted arcs efficiently, with the assistance of LLM tools streamlining the correction process.

\section{\uppercase{Conclusions and Future Work}}
\label{sec:conclusion}

This project tackled the challenge of analyzing the interwoven narrative arcs of serialized television. The multi-agent system introduced represents a step toward automating the traditionally manual process of identifying and organizing narrative arcs for deeper study. 

While the results are promising, they also reveal limitations, particularly the reliance on textual paratexts, such as episode summaries, as the primary input.
Summaries focus on major plot-driving events, neglecting subtle details found in visuals, dialogue, or other storytelling layers. Consequently, the system occasionally misses overlapping arcs or blends distinct storylines.

However, this limitation also presents an opportunity. The system is particularly well-suited for serialized written narratives, where all story elements are text-based. Serialized novels, episodic web fiction, and other text-driven storytelling formats eliminate the multimodal challenges of TV series, offering a clearer field for arc extraction and progression mapping. With minor adjustments, the workflow can be adapted to handle chapters, segments, or semantic chunks of text \cite{qu2024semantic}, treating them as "episodes" to track storylines over an extended series.

Future work will enhance the system by integrating richer inputs such as subtitles, scene descriptions, and multimodal data. Refinements to the multi-agent system are also needed to improve reliability, particularly in managing overlapping arcs and deduplication. Expanding testing across diverse genres will further refine the system’s versatility and applicability.

Ultimately, the combination of automation and human oversight remains indispensable. While machines excel at processing and organizing large datasets, human interpretation continues to provide the understanding required for meaningful narrative analysis.

\bibliographystyle{apalike}
{\small
\bibliography{PAPER}}

\end{document}